# Machine Common Sense
# Concept Paper

David Gunning
DARPA/I2O
October 14, 2018

## Introduction

This paper summarizes some of the technical background, research ideas, and possible development strategies for achieving machine common sense. This concept paper is not a solicitation and is provided for informational purposes only. The concepts are organized and described in terms of a modified set of Heilmeier Catechism questions.

## What are you trying to do?

Machine common sense has long been a critical—but missing—component of Artificial Intelligence (AI). Recent advances in machine learning have resulted in new AI capabilities, but in all of these applications, machine reasoning is narrow and highly specialized. Developers must carefully train or program systems for every situation. General commonsense reasoning remains elusive.

Wikipedia defines common sense as, the basic ability to perceive, understand, and judge things that are shared by ("common to") nearly all people and can reasonably be expected of nearly all people without need for debate. It is common sense that helps us quickly answer the question, "can an elephant fit through the doorway?" or understand the statement, "I saw the Grand Canyon flying to New York." The vast majority of common sense is typically not expressed by humans because there is no need to state the obvious. We are usually not conscious of the vast sea of commonsense assumptions that underlie every statement and every action. This unstated background knowledge includes: a general understanding of how the physical world works (i.e., intuitive physics); a basic understanding of human motives and behaviors (i.e., intuitive psychology); and knowledge of the common facts that an average adult possesses. Machines lack this basic background knowledge that all humans share. The obscure-but-pervasive nature of common sense makes it difficult to articulate and encode in machines.

The absence of common sense prevents intelligent systems from understanding their world, behaving reasonably in unforeseen situations, communicating naturally with people, and learning from new experiences. Its absence is perhaps the most significant barrier between the narrowly focused AI applications we have today and the more general, human-like AI systems we would like to build in the future.

Machine common sense remains a broad, potentially unbounded problem in AI. There are a wide range of strategies that could be employed to make progress on this difficult challenge. This paper discusses two diverse strategies for focusing development on two different machine commonsense services:

- A service that learns from experience, like a child, to construct computational models that mimic the core domains of child cognition for objects (intuitive physics), agents (intentional actors), and places (spatial navigation); and

- A service that learns from reading the Web, like a research librarian, to construct a commonsense knowledge repository capable of answering natural language and image-based questions about commonsense phenomena.

## If you are successful, what difference will it make?

If successful, the development of a machine commonsense service could accelerate the development of AI for both defense and commercial applications. Here are four broad uses cases that apply to single AI applications, symbiotic human-machine partnerships, and fully autonomous systems:

- *Sensemaking* – any AI system that needs to analyze and interpret sensor or data input could benefit from a machine commonsense service to help it interpret and understand real world situations;
- *Monitoring the reasonableness of machine actions* – a machine commonsense service would provide the ability to monitor and check the reasonableness (and safety) of any AI system's actions and decisions, especially in novel situations;
- *Human-machine collaboration* – all human communication and understanding of the world assumes a background of common sense. A service that provides machines with a basic level of human-like common sense would enable them to more effectively communicate and collaborate with their human partners, and;
- *Transfer learning (adapting to new situations)* – a package of reusable commonsense knowledge would provide a foundation for AI systems to learn new domains and adapt to new situations without voluminous specialized training or programming.

## How is it done today? What are the limitations of current practice?

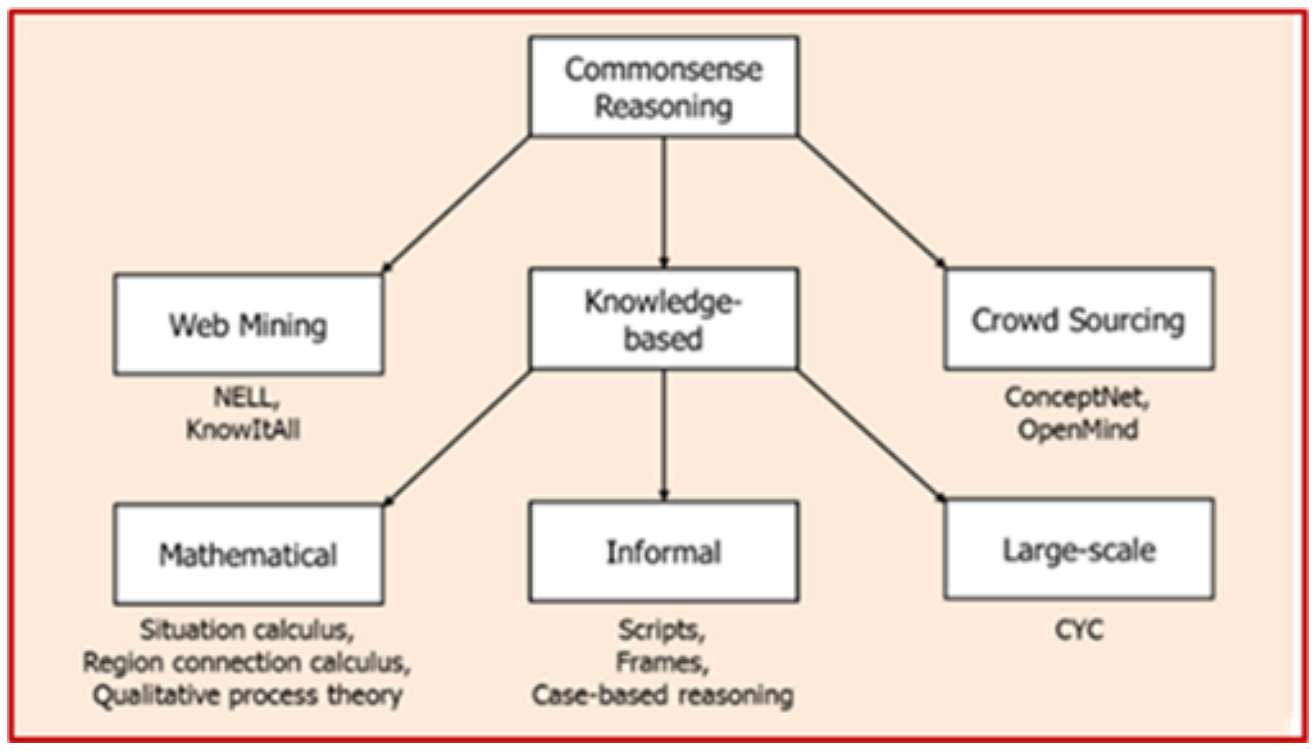

*Figure 1: Taxonomy of Approaches to Commonsense Reasoning [1]*

A 2015 survey of commonsense reasoning in AI summarized the major approaches taken in the past [1], including the taxonomy of approaches shown in Figure 1 below. Shortly after co-founding the field of AI in the 1950's, John McCarthy speculated that *programs with common sense* could be developed using formal logic [2]. This suggestion led to a variety of efforts to develop logic-based approaches to commonsense reasoning (e.g., situation calculus [3], naïve physics [4], default reasoning [5], non-monotonic logics [6], description logics [7], and qualitative reasoning [8]), less formal knowledge-based approaches (e.g., frames [9], and scripts [10]), and a number of efforts to create logic-based ontologies (e.g., WordNet [11], VerbNet [12], SUMO [13], YAGO [14], DOLCE [15], and hundreds of smaller ontologies on the Semantic Web [16]).

The most notable example of this knowledge-based approach is Cyc [17], a 35-year effort to codify common sense into an integrated, logic-based system. The Cyc effort is impressive. It covers large areas of common sense knowledge and integrates sophisticated, logic-based reasoning techniques. Figure 2 illustrates the concepts covered in Cyc's extensive ontology. Yet, Cyc has not achieved the goal of providing a generally useful commonsense service. There are many reasons for this, but the primary one is that Cyc, like all of the knowledge-based approaches, suffers from the brittleness of symbolic logic. Concepts are defined in black or white symbols, which never quite match the subtleties of the human concepts they are intended to represent. Similarly, natural language queries never quite match the precise symbolic concepts in Cyc. Cyc's general ontologies always need to be tailored and refined to fit specific applications. When combined into large, handcrafted systems such as Cyc, these symbolic concepts yield a complexity that is difficult for developers to understand and use [18].

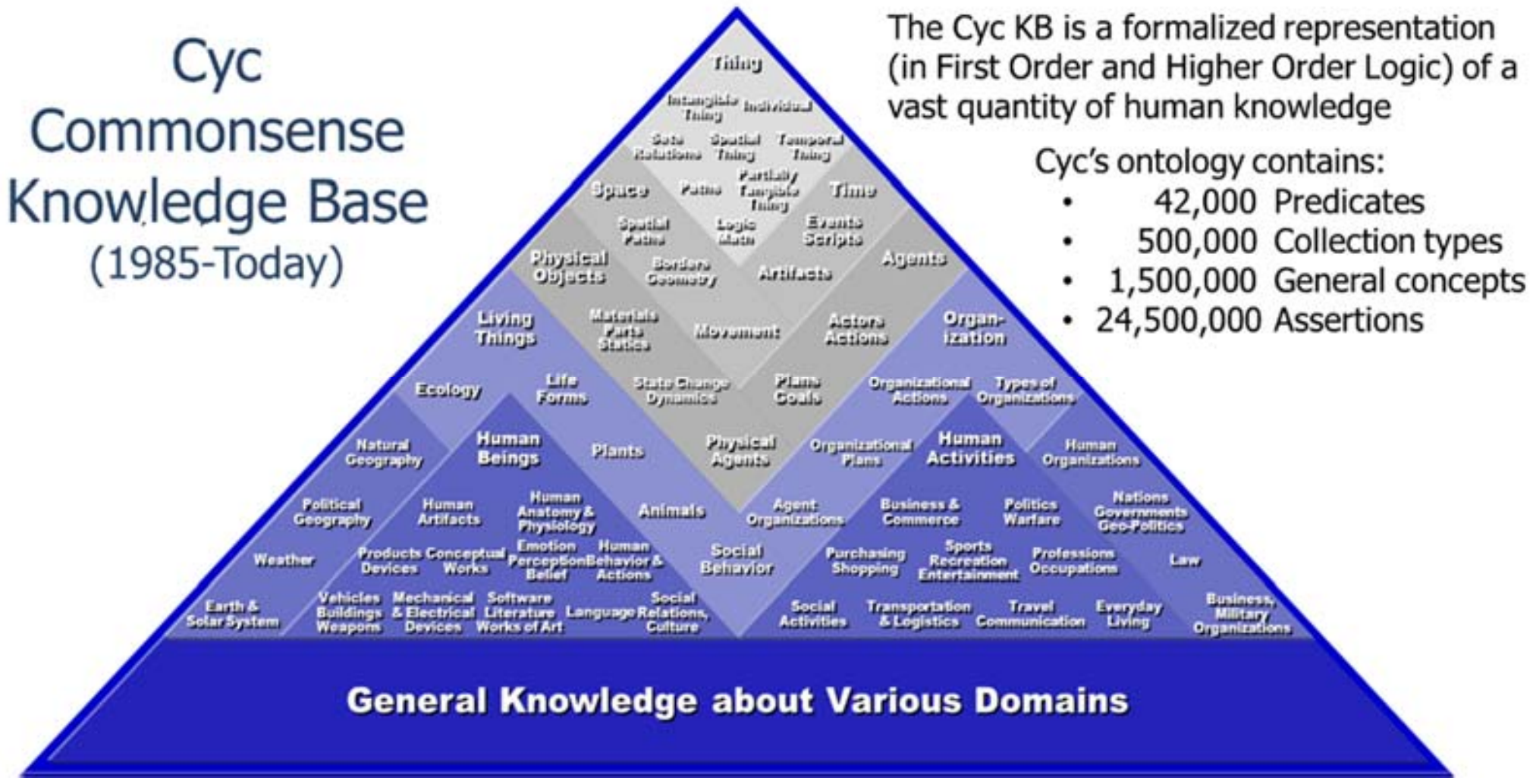

*Figure 2: Cyc Knowledge Base [17]*

More recently, as machine learning and crowdsourcing have come to dominate AI, those techniques have also been used to extract and collect commonsense knowledge from the Web. Several efforts have used machine learning and statistical techniques for large-scale information extraction from the entire Web (e.g., KnowItAll [19]) or from a subset of the Web such as Wikipedia (e.g., DBPedia [20]). Several other systems have used crowdsourcing to acquire knowledge from the general public via the Web, such as OpenMind [21] and ConceptNet [22].

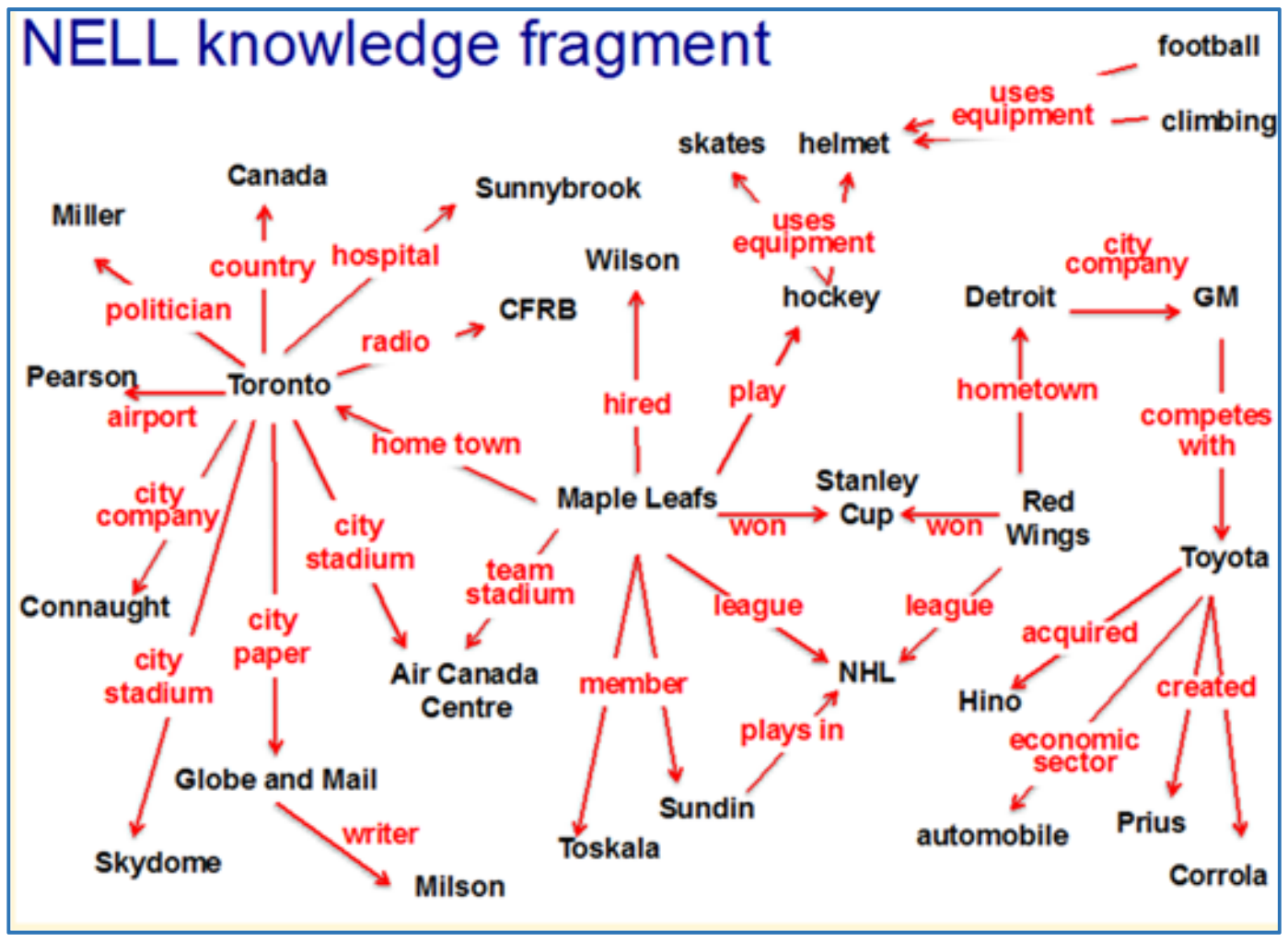

*Figure 3: NELL Knowledge Fragment [24]*

The most notable and comprehensive example that combines machine leaning with crowdsourcing is Tom Mitchell's Never Ending Language Learning (NELL) system [23][24]. NELL has been learning to read the Web 24 hours a day since January 2010. So far, NELL has acquired a knowledge base with 120 million diverse, confidence-weighted beliefs (e.g., won(MapleLeafs, StanleyCup)), as shown in Figure 3. The inputs to NELL include an initial seed ontology defining hundreds of categories and relations that NELL is expected to read about, and 10 to 15 seed examples of each category and relation. Given these inputs and access to the Web, NELL runs continuously to extract new instances of categories and relations. NELL also uses crowdsourcing to provide feedback from humans in order to improve the quality of its extractions. Although machine learning approaches like NELL are much more scalable (as opposed to hand-coded symbolic engineering approaches) at accumulating large amounts of knowledge, their relatively shallow semantic representations suffer from ambiguities and inconsistencies. While approaches like NELL continue to make significant progress, they generally lack sufficient semantic understanding to enable reasoning beyond simple answer lookup. These approaches have also fallen short of producing a widely useful commonsense capability. Machine common sense remains an unsolved problem.

One of the most critical—if not THE most critical—limitation has been the lack of flexible, perceptually grounded concept representations, like those found in human cognition. There is significant evidence from cognitive psychology and neuroscience to support the Theory of Grounded Cognition [25][26], which argues that concepts in the human brain are grounded in perceptual-motor memories and experiences. For example, if you think of the concept of a *door*, your mind is likely to imagine a door you open often, including a mild activation of the neurons in your arm that open that door. This grounding includes perceptual-motor simulations that are used to plan and execute the action of opening the door. If you think about an abstract metaphor, such as, "when one door closes, another opens," some trace of that perceptual-motor experience is activated and enables you to understand the meaning of that abstract idea. This theory also argues that much of human common sense occurs through mental simulation using these perceptual-motor concepts. For example, if you are asked, "Can an elephant fit through the doorway?", your mind is likely to run a quick perceptual simulation to answer the question.

Linguists, such as George Lakoff, argue that perceptually grounded concepts are the key to understanding metaphor, and metaphor is the key to understanding human thought [27][28][29]. Discovering the right grounding is critical for both learning commonsense concepts and performing commonsense reasoning. Although there is no general agreement on the importance of grounded cognition and metaphor in AI, it seems clear that development of more perceptually grounded representations will be critical for making progress on machine common sense, where matching human concept representations is critical. Such representations would not only get us closer to human cognition, they may also be the key to integrating machine learning and machine reasoning.

## What is new in your approach and why do you think it will be successful?

There has been significant progress in AI along a number of dimensions that make it possible to address this difficult problem now. There continues to be rapid advancement in all aspects of machine learning, especially deep learning, that is producing new representations and new techniques for semi-supervised, self-supervised, and unsupervised learning. This progress has created a resurgence of young researchers who are using these new representations and techniques to take on the common sense problem. They have produced four areas of new research, in particular, that answer the question, "why now?": (1) learning grounded representations; (2) learning commonsense knowledge from the Web; (3) learning predictive models from experience; and (4) understanding and modeling childhood cognition.

### Learning Grounded Representations

One of the most useful by-products of deep learning has been the use of *embeddings* to represent semantic concepts. Word embeddings, such as Word2Vec [30][31], are now widely used in natural language processing to map word phrases to vectors of real numbers. An embedding typically transforms the representation of words from a space with one dimension per word, to a continuous vector space with less dimensionality. Neural networks are often used to learn these embeddings to represent semantic similarities between words, based on the statistics of neighboring words in large samples of natural language data. Words with similar meanings are close together in the embedding space. Google reports that their multilingual neural machine translation system is able to use embeddings, learned from translating multiple language pairs, as a kind of Interlingua, to perform zero-shot translation between two languages – without specific training for that language pair [32].

More generally, semantic concepts from any source (language, vision, auditory, or motor) can be learned and represented in this type of vector-based embedding space. Embeddings are widely used (by all of the researchers cited here and many others) to learn perceptually grounded representations from language, images, and video, as well as simulated and real environments. These representations are not perfect and have limitations. Researchers are actively trying to discover new techniques to effectively compose, simulate, and reason with these representations. In addition, other researchers have developed promising alternative (non-deep learning) representations. For example: Josh Tenenbaum (MIT) and his colleagues have developed rich probabilistic representations that mimic human learning [33][34]; and Song-Chun Zhu (UCLA) has developed an array of techniques based on stochastic and-or-graphs [35][36][37]. All of these new representations show promise as a better foundation for learning human-like common sense concepts.

## Learning Commonsense Knowledge from the Web

Much of the new work focuses on learning commonsense knowledge from images and language on the Web. For example, Abhinav Gupta, a recent addition to the CMU faculty, has created a companion to NELL, the Never Ending Image Learning (NEIL) system, that uses semi-supervised, deep learning algorithms to discover commonsense relationships (e.g., "Corolla is a kind of a Car" and "Wheel is a part of Car") from images on the Web [38]. Yejin Choi, a new faculty member at UW, has led a series of projects to learn commonsense knowledge from language on the Web (e.g., verb physics [39], event inferences [40], story understanding [41]).

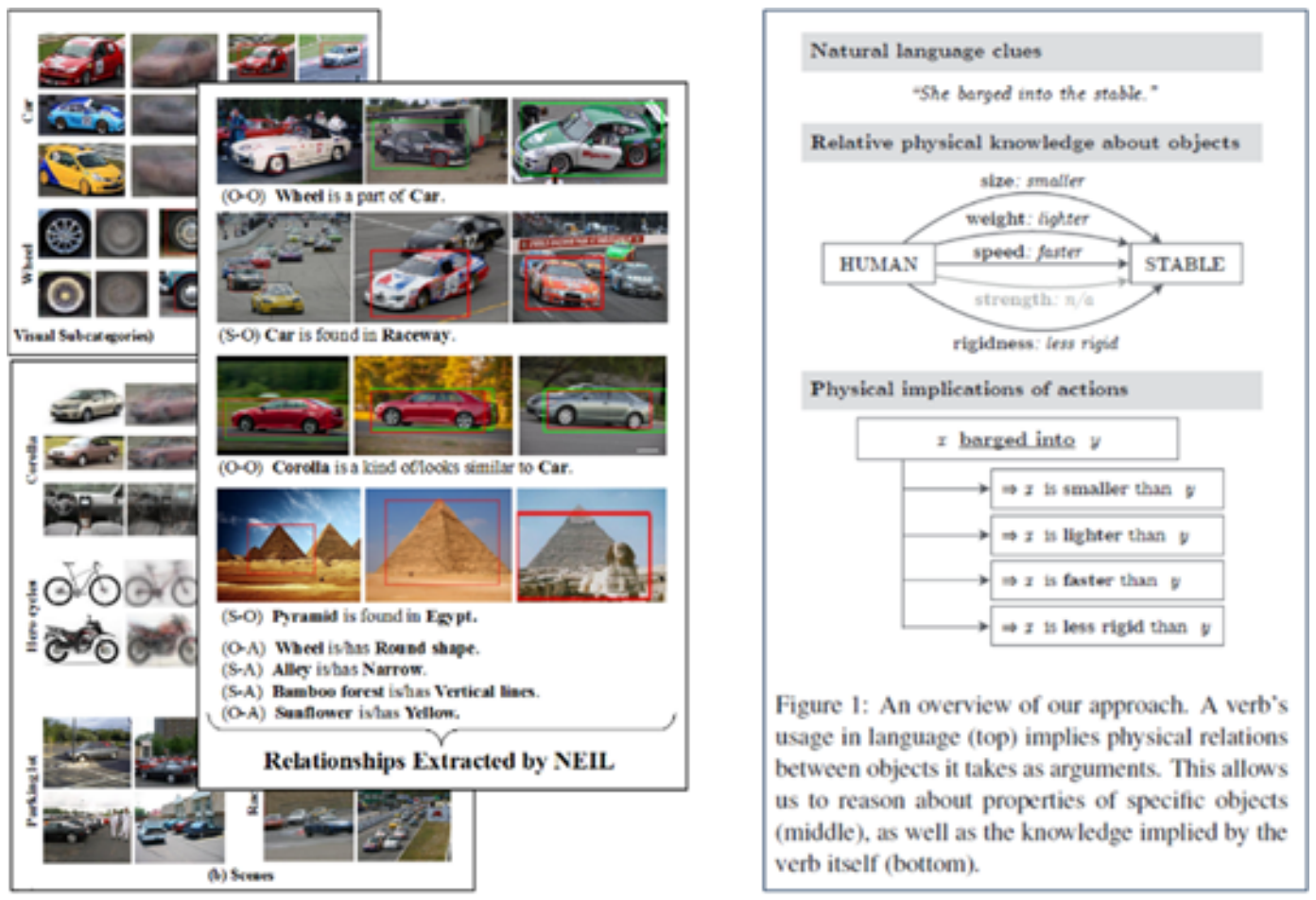

NEIL: Extracting Visual Knowledge from Web Data

VERB PHYSICS: Relative Physical Knowledge of Actions and Objects

*Figure 4: Examples of Learning Commonsense Knowledge from Images (NEIL [38]) and Language (VERB PHYSICS [39])*

These researchers are discovering new techniques for extracting commonsense knowledge from language [42][43][44][45], vision [46][47][48][49], and robotics [50][51][52]. Others have used techniques such as knowledge-based completion [53][54]. These researchers include rising stars in the DARPA community, including: Mohit Bansel, a 2018 Young Faculty Award winner from UNC; Xiao Lin, a D60 Riser from (SRI); and Stefan Lee, a D60 Riser from (GA Tech). Moreover, cutting edge research in deep learning is going well beyond supervised classification to create more complete systems capable of memory [55], 'mental' simulation [56], and multi-step reasoning [57].

## Learning Predictive Models from Experience

Researchers have also discovered how to use vector-based embeddings to learn predictive models of commonsense phenomenon from videos and simulations. A landmark paper published in 2016 demonstrated that self-supervised techniques could learn predictive models from video by learning to predict changes in these internal, embedded representations [58]. The basic idea is to train a deep network to predict the next event in an unlabeled video sequence. No hand labeling is needed as the ground truth appears in future frames. Previous work had tried to predict events at the pixel level, which proved too difficult. This research demonstrated it was possible to learn predictive models of everyday events by predicting changes in the feature space of the deep learning system (Figure 5).

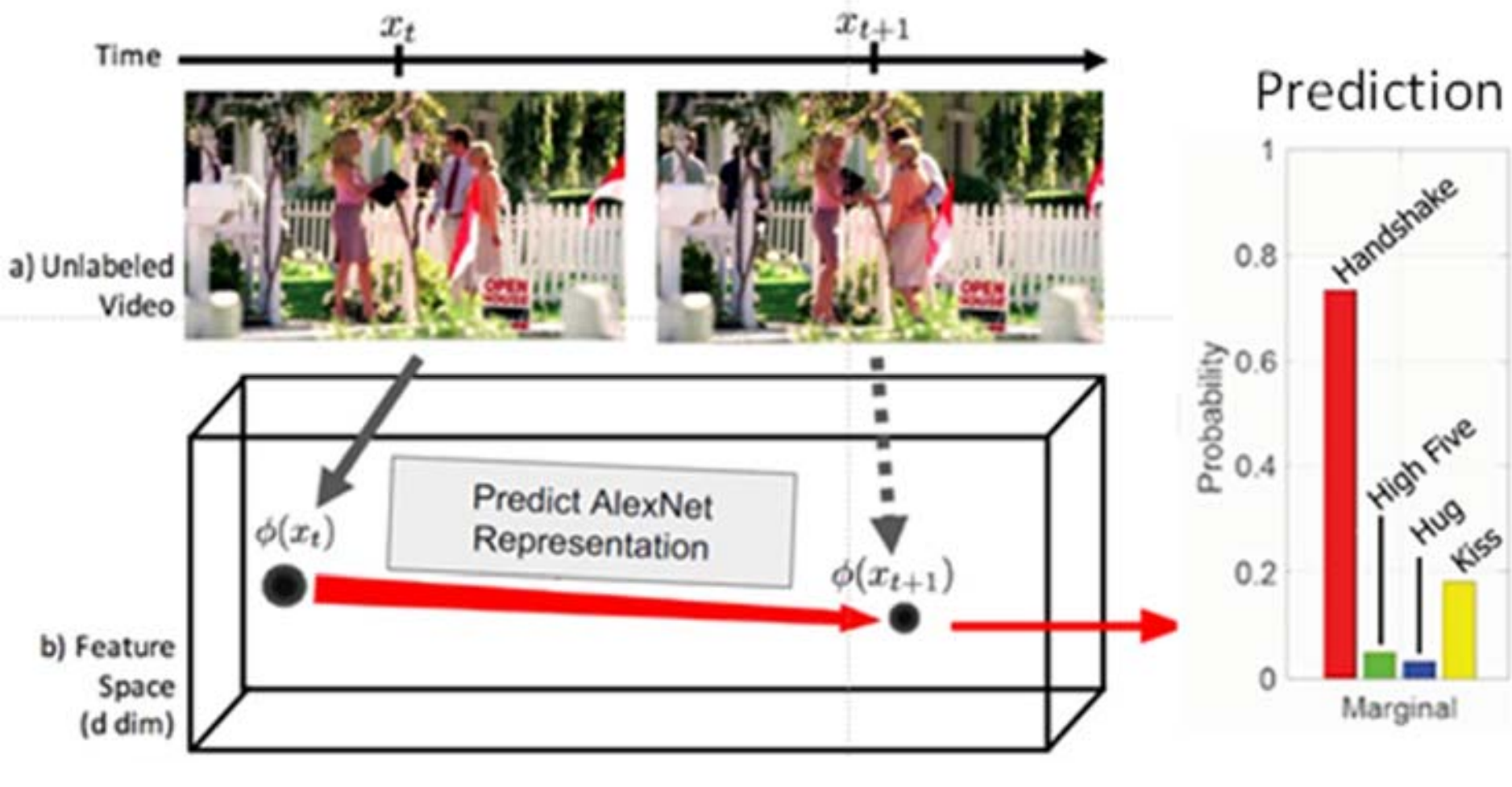

*Figure 5: Anticipating Visual Representations from Unlabeled Video [58]*

This self-supervised technique is now widely used in deep learning research to learn predictive models from video, simulation, and real world activities. For example, Facebook researchers have used this technique to learn an intuitive physics model of block towers [59]. Using both physical blocks and ones in a simulated 3D game engine, they created small towers of blocks whose stability was randomized and then rendered collapsing (or remaining upright) into a video (Figure 6). The researchers then trained a deep learning system, by watching these videos of the simulated and real environments, to accurately predict the outcomes, as well as estimate block trajectories. The deep learning system then used this self-supervised technique to learn a predictive model of this simple physics phenomenon. The promise of these techniques has prompted Yann LeCun (Facebook) to propose that extensions of deep learning could now be used to learn predictive models of commonsense reasoning by “replacing symbols with vectors and replacing reasoning with algebra” [60].

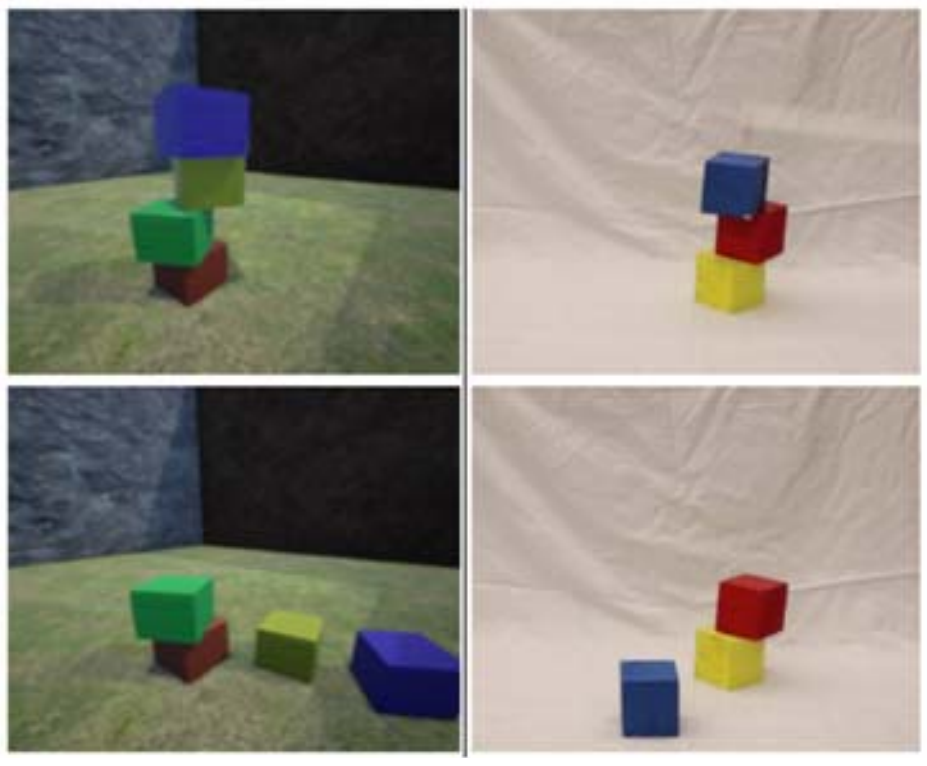

*Figure 6: Block Tower Examples [59]*

## Understanding and Modeling Childhood Cognition

Researchers who study childhood cognition now have years of experimental results that allow them to map out the cognitive capacities of children. The field of cognitive development is at a point where it can provide empirical and theoretical guidance for building intelligent machines that think and learn like children. In particular, developmental psychologists have intensively studied children's knowledge in six domains (Table 1). Some believe that each of these domains constitutes a distinct and relatively autonomous system of knowledge, an idea that has been codified in the Theory of Core Knowledge. Others believe that these domains interact from the beginning of life. Developmental psychologists agree, however, that abilities to reason about objects, agents, places, number, geometry, and the social world, as described in the Theory of Core Knowledge, emerge early and serve as crucial foundations for later learning [61][62][63]:

*Table 1: Theory of Core Knowledge*

| Domain | Description |
|---|---|
| Objects | supports reasoning about objects and the laws of physics that govern them |
| Agents | supports reasoning about agents that act autonomously to pursue goals |
| Places | supports navigation and spatial reasoning around an environment |
| Number | supports reasoning about quantity and how many things are present |
| Forms | supports representation of shapes and their affordances |
| Social Beings | supports reasoning about Theory of Mind and social interactions |

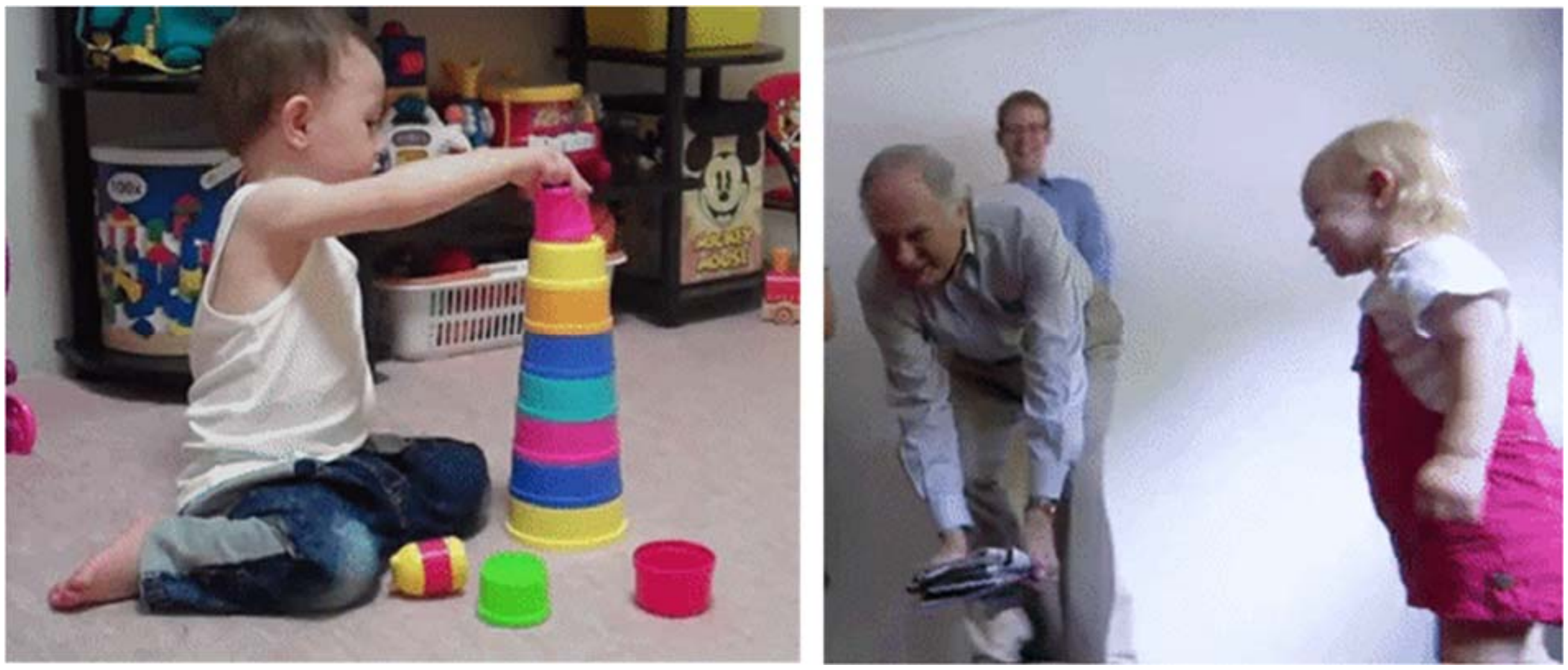

*Figure 7: Child Cognition for Objects (left) and Agents (right) [Source: medium.com]*

These core domains serve as the fundamental building blocks of human intelligence and common sense, especially the core domains of objects (intuitive physics), agents (intentional actors), and places (spatial navigation). For example, the core domain of objects not only provides the fundamental concepts for understanding the physical world, but also provides the foundation for understanding causality. The core domain of agents not only provides the fundamental concepts for understanding intentional actors and Theory of Mind (TOM), but also provides the foundation for dealing with the “frame problem” in AI (i.e., knowing that objects in a scene only change if acted on by an agent). The core domain of places not only provides the fundamental concepts for navigation, but also provides the foundation for spatial memory and spatial reasoning.

Each core domain is characterized by key principles and signature limits. The object domain, for example, is characterized by three key principles that guide reasoning in that domain:

- The Cohesion Principle – objects should hold together across time and space;
- The Continuity Principle – objects should move along continuous paths in time and space; and
- The Contact Principle – objects should only move with contact from another object.

Children expect objects to behave according to each domain's principles and are surprised when those principles are violated (i.e., Violation of Expectation (VOE)). A child's surprise has become a primary means of studying child cognitive abilities and is widely used as an experimental measure to study the precise development of these six domains, even in pre-lingual children. For example, the MIT Early Childhood Lab has developed the *LookIt* test environment that enables them to conduct crowdsourced studies of child cognition, over the Web. In one of their current studies, "Your baby, the physicist," children between 4-12 months can view a 15-minute video that tests their physics knowledge. By recording facial expressions using a webcam, researchers are able to determine which physics principles match or violate the child's expectations.

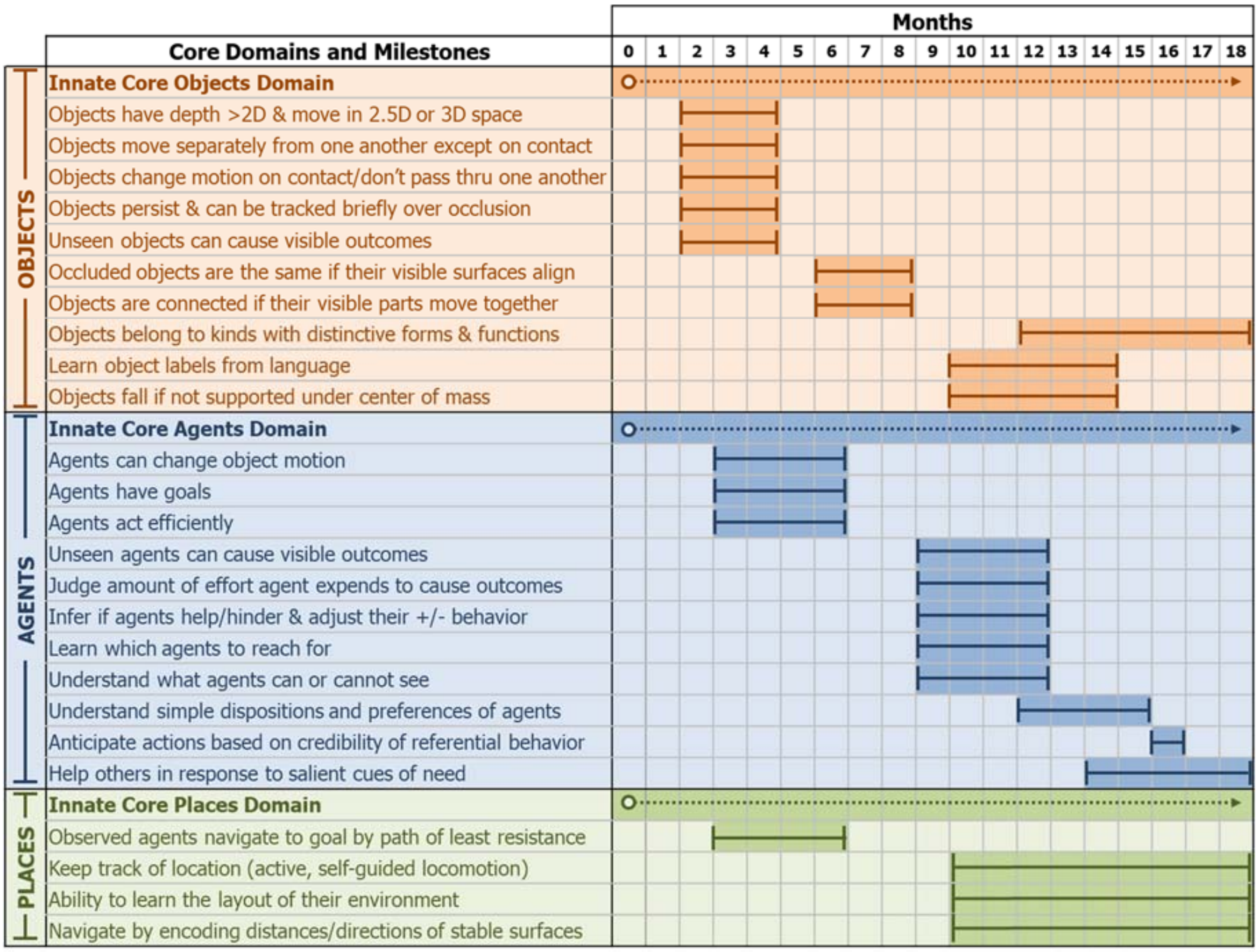

*Figure 8: Cognitive Development Milestones (0-18 months)*

As a result of these new experimental techniques, developmental psychologists are now able to map the cognitive capacities of children. Figure 8 illustrates key stages in the current understanding of the developmental sequence for the three core domains of objects, agents, and places for children from 0 to 18 months. This sequence provides an excellent set of target milestones for AI researchers to mimic as a strategy for developing a new foundation for machine common sense. While these milestones are particularly useful, these are just a selection of those the literature suggests. In addition, research in development is ongoing and it is helpful to consider Figure 8 as including “error bars” on both the columns (time of acquisition) and rows (the conceptual split and grouping of the abilities and understandings of children).

AI researchers have begun to use these results from developmental psychology to create computational models of child cognition. Josh Tenenbaum (MIT) has used this work from cognitive psychology to develop probabilistic models of human-like learning, including computational models of intuitive physics that mimic child cognition [64]. Figure 9 shows an example of probabilistic predictions made by this intuitive physics engine.

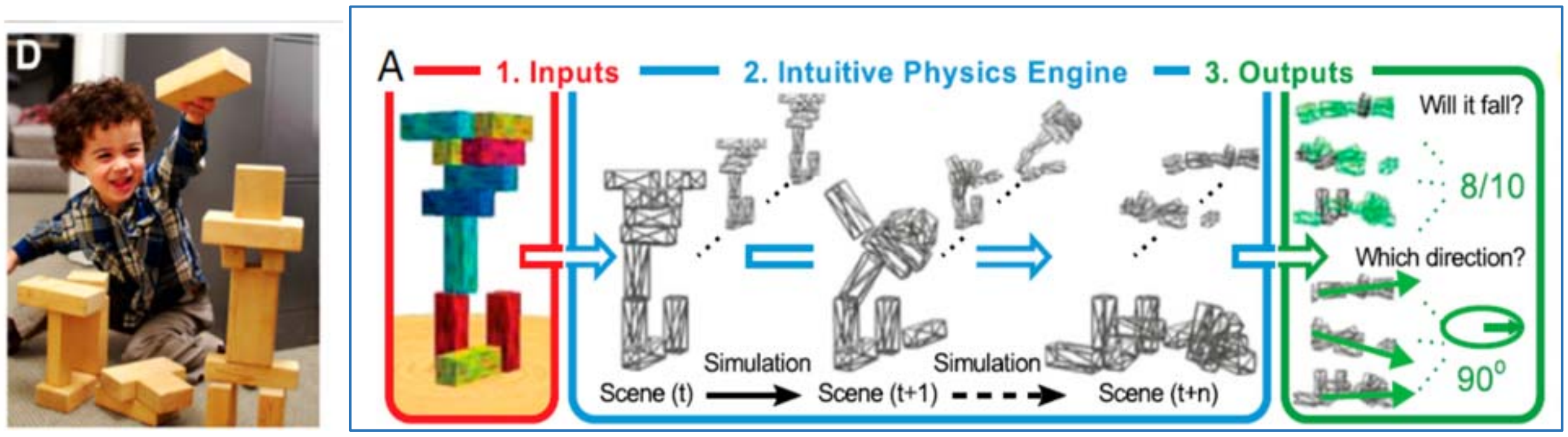

*Figure 9: Intuitive Physics Engine [64]*

Researchers at DeepMind have also trained deep learning models of intuitive physics by watching video renderings of simple blocks world simulations. Moreover, they demonstrated a scheme for using the same VOE method used in developmental psychology to evaluate how well the artificial models mimic child cognition [65].

In summary, general progress in AI, as well as the specific progress in learning grounded representations, learning commonsense knowledge from the Web, learning predictive models from experience, and understanding and modeling childhood cognition, presents interesting opportunities for achieving machine common sense.

## What are the mid-term and final “exams” to check for success?

The potential strategies discussed would develop two different commonsense services, each with their own evaluation method:

- Foundations of Human Common Sense: a service that learns from experience, like a child, to construct computational models that mimic the core knowledge systems of cognition for objects (intuitive physics), places (spatial navigation), and agents (intentional actors). These models would be evaluated against the cognitive development milestones as evidenced in developmental psychology experiments with children from 0-18 months old, as show in Figure 8 above.

- Broad Common Knowledge: a service that learns from reading the Web, like a research librarian, to construct a commonsense knowledge repository capable of answering natural language and image-based queries about commonsense phenomena. This service would attempt to mimic the general knowledge of an average adult, as measured by the Allen Institute for Artificial Intelligence (AI2) Common Sense benchmark tests.

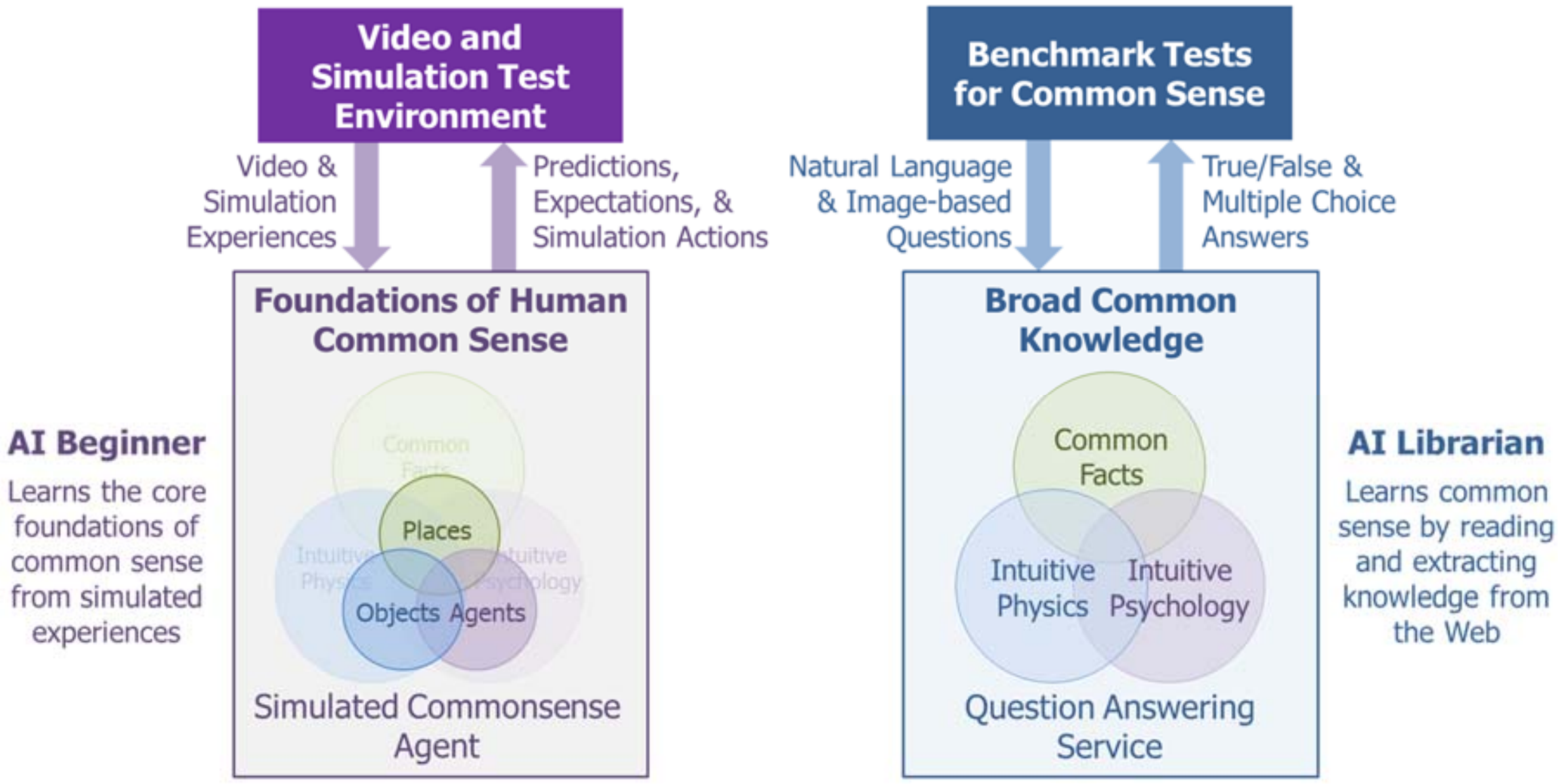

*Figure 10: Possible Machine Common Sense Services*

## Foundations of Human Common Sense

One strategy for developing a commonsense service would be to design and construct computational models that mimic the cognitive capabilities of children, 0-18 months old, for the three core domains of objects, agents, and places. A variety of strategies could achieve this goal, ranging from pre-building initial models to learning everything from scratch, using any combination of symbolic, probabilistic, or deep learning techniques. It is expected that these computational models would need some form of perceptually grounded representations, combined with reasoning and simulation methods that work with those representations.

A key component of such a strategy is likely require the consolidation, refinement, and extension of the psychological theories. Both AI and developmental psychology expertise would be needed to produce both computational models and refined psychological theories of child cognition. Both might benefit from companion research experiments in developmental psychology to answer critical design questions relevant to the computational models, and (possibly) to test predictions made by the models through supplemental research with children.

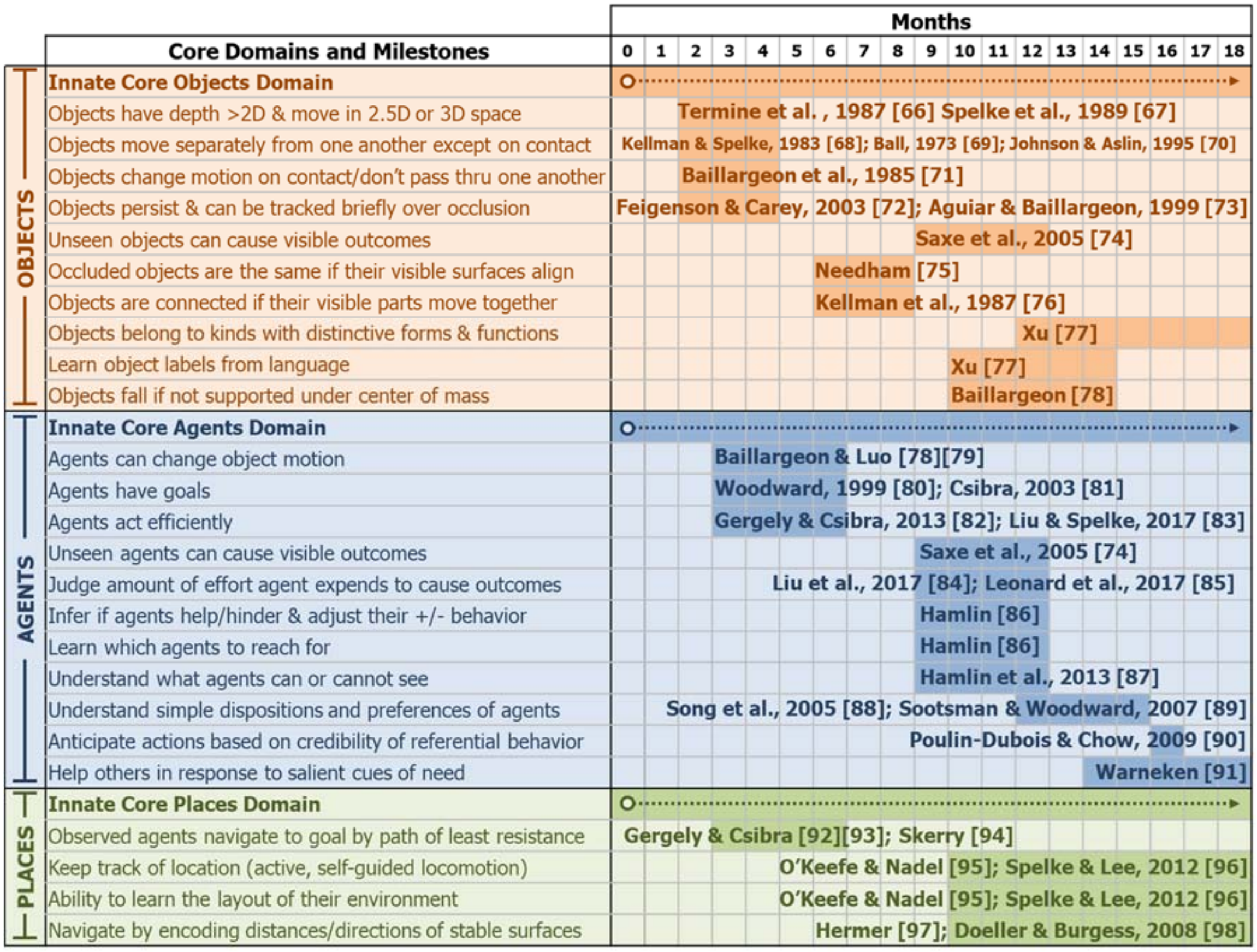

*Figure 11: Research on Cognitive Development Milestones (0-18 months)*

The computational models could be evaluated against the cognitive development milestones as evidenced in developmental psychology experiments with children from 0-18 months old. Figure 11 lists examples of the research supporting each of the milestones. The body of research could be used to construct specific test problems for each milestone to evaluate the computational models at three levels of performance:

- Prediction/expectation: the test environment will present the computational models with videos and simulation experiences of the type used to test child cognition for each cognitive milestone. The models will produce a prediction or expectation output that will be used to determine if the model matches human cognitive performance. The models will provide a measurable VOE signal when shown a possible next event, for direct comparison to the VOE results observed in children.
- Experience learning: the test environment will present the computational models with videos and simulation experiences in which a new object, agent, or place is introduced. The models will be tested to determine that they are able to learn the properties of the newly introduced item in a way that matches human cognitive performance.
- Problem solving: the test environment will present the computational models with videos and simulation experiences in which a problem solving task is introduced. The models will be tested to determine they solve the problem in a way that matches human cognitive performance.

Evaluation of the computational models would require:

- a test infrastructure consisting of a library of videos and a high fidelity 3D simulation environment (examples of existing 3D simulation environments are shown in Figure 12 below);
- development of specific test problems, based on the results of developmental psychology experiments on child cognition (such as the examples shown in Figure 11 above), to evaluate the computational models at various levels of performance.

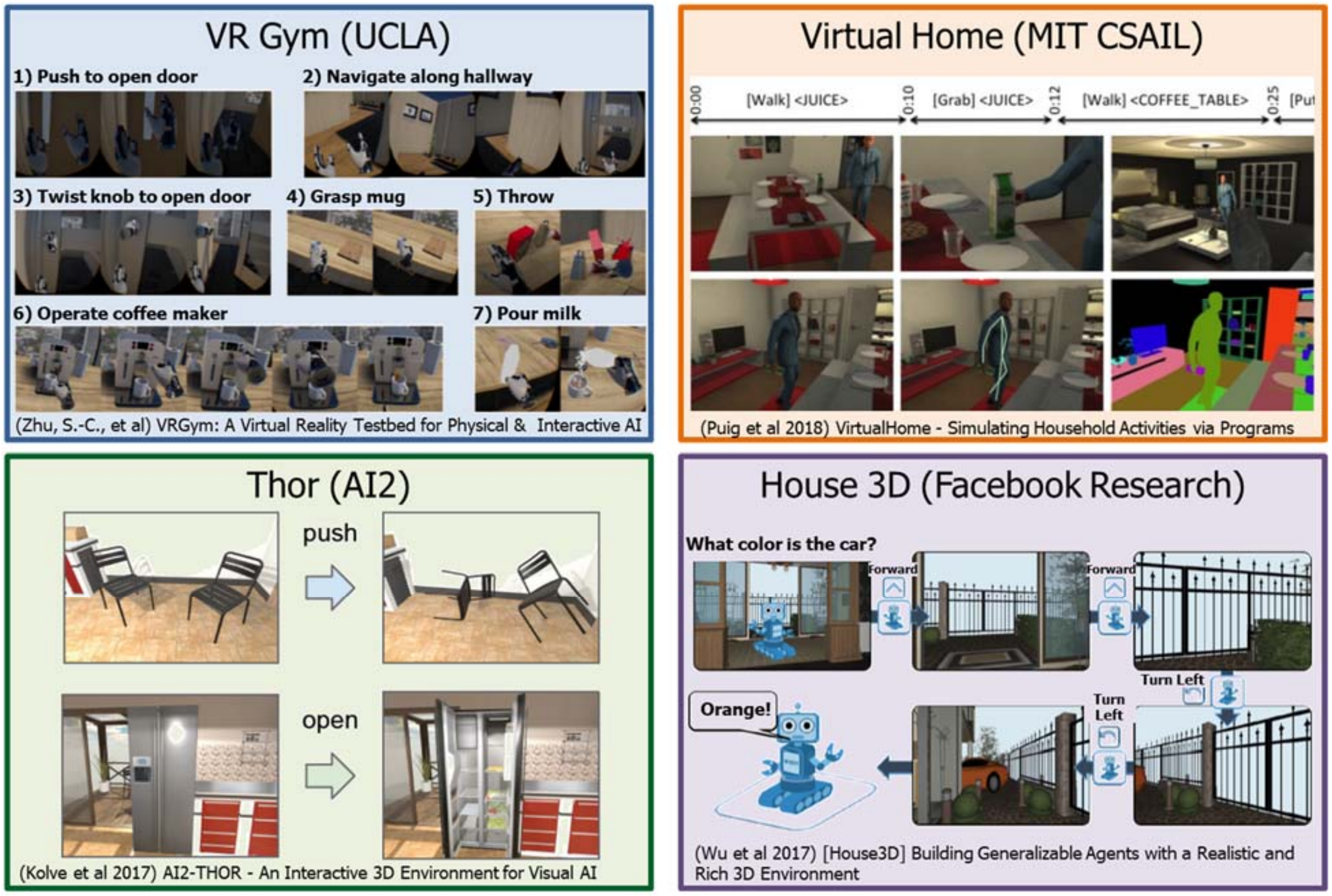

*Figure 12: Examples of 3D Simulation Environments*

## Broad Common Knowledge

Another strategy for developing a commonsense service would be to learn/extract/construct a commonsense knowledge repository capable of answering natural language and image-based questions about commonsense phenomena, such as those from the AI2 Benchmarks for Common Sense. (https://allenai.org/commonsense/). A variety of strategies could be used to construct a repository of broad common knowledge, including any combination of manual construction, information extraction, machine learning, and crowdsourcing techniques. Techniques could be artificial or biologically inspired.

A broad common knowledge service could be evaluated against established benchmarks for common sense. AI2 has developed novel crowdsourcing techniques to generate a massive corpus of common sense test questions [99]. AI2 has also developed a sequestered, automated test environment, automated scoring algorithms, and a leaderboard to publish results. Such benchmarks would measure the performance of a question answering (QA) service for natural language inference (NLI), NLI combined with vision, abductive NLI, physical interaction QA, social interaction QA, and others (Figure 13).

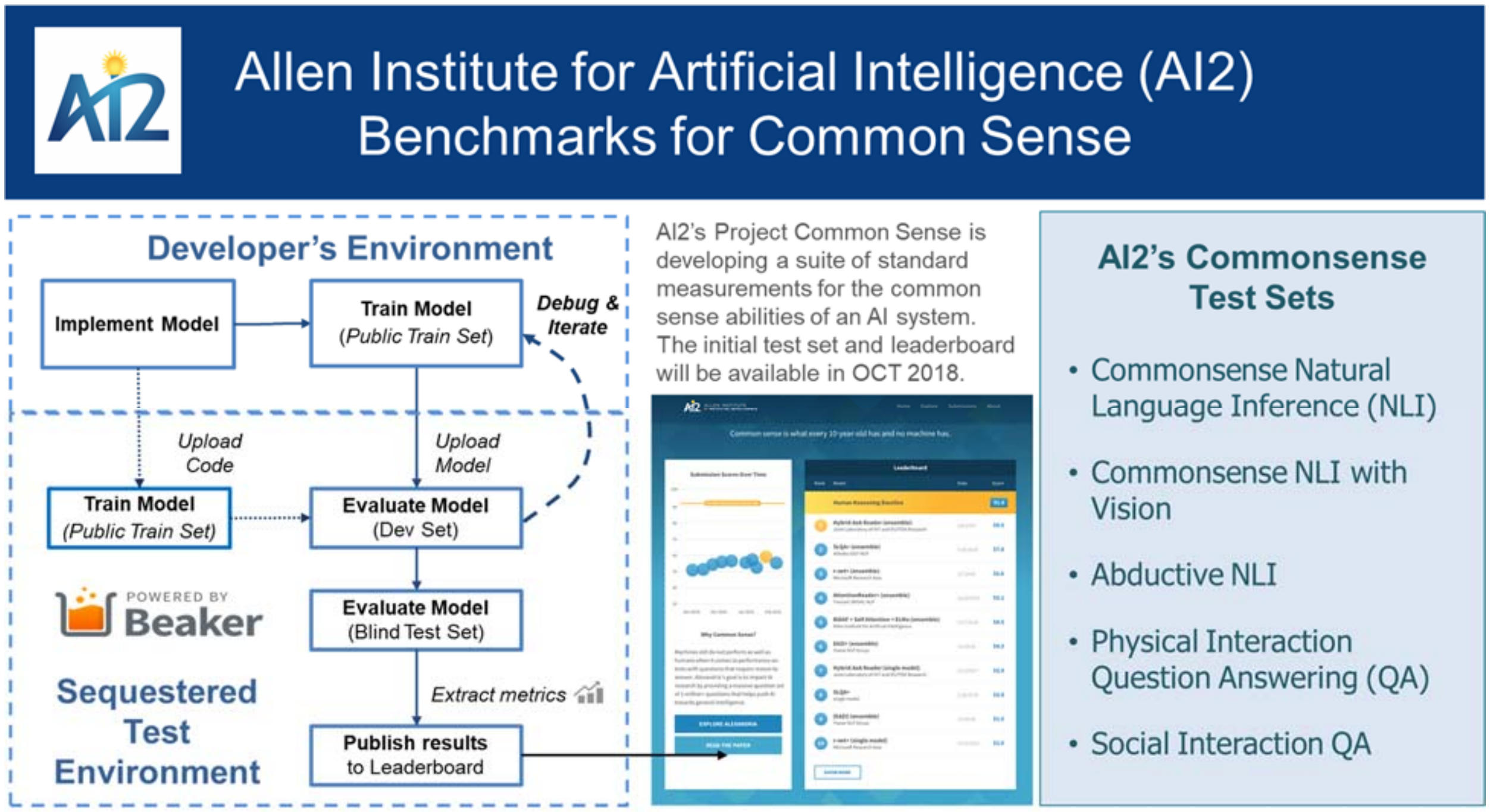

*Figure 13: AI2 Benchmarks for Common Sense [source: AI2]*

## Acknowledgements

The author thanks: Murray Burke for his sage advice and long-standing expertise in AI; Dr. Joshua Alspector for his expertise in deep learning and insights into its potential for achieving machine common sense; and Ms. Marisa Carrera for her exceptional technical support and editorial skills.